\documentclass[10pt,twocolumn,letterpaper]{article}

\pdfoutput=1

\usepackage{arxiv}
\usepackage{graphicx}
\usepackage{amsmath}
\usepackage{amssymb}
\usepackage{booktabs}

\usepackage{array}
\usepackage{textcomp}
\usepackage{stfloats}
\usepackage{url}
\usepackage{verbatim}
\usepackage{graphicx}
\usepackage{url}

\usepackage{multirow}
\usepackage{times}
\usepackage{epsfig}
\usepackage{graphicx}
\usepackage{amssymb}
\usepackage{adjustbox}

\usepackage{amssymb}

\usepackage[normalem]{ulem}
\useunder{\uline}{\ul}{}

\usepackage{xspace}

\makeatletter
\DeclareRobustCommand\onedot{\futurelet\@let@token\@onedot}
\def\@onedot{\ifx\@let@token.\else.\null\fi\xspace}

\def\eg{\emph{e.g}\onedot}

\def\etal{\emph{et al}\onedot}
\makeatother

\iccvfinalcopy 

\ificcvfinal\fi

\begin{document}

\title{Sketch-an-Anchor: Sub-epoch Fast Model Adaptation for Zero-shot Sketch-based Image Retrieval}

\author{Leo Sampaio Ferraz Ribeiro\\
ICMC, Universidade de São Paulo\\
São Carlos/SP, Brazil\\
{\tt\small leo.sampaio.ferraz.ribeiro@alumni.usp.br}
\and
Moacir Antonelli Ponti\\
Mercado Libre\\
Osasco/SP, Brazil\\
{\tt\small moacir.ponti@mercadolivre.com}
}
\maketitle

\ificcvfinal\thispagestyle{empty}\fi

\begin{abstract}
   Sketch-an-Anchor is a novel method to train state-of-the-art Zero-shot Sketch-based Image Retrieval (ZSSBIR) models in under an epoch. Most studies break down the problem of ZSSBIR into two parts: domain alignment between images and sketches, inherited from SBIR, and generalization to unseen data, inherent to the zero-shot protocol. We argue one of these problems can be considerably simplified and re-frame the ZSSBIR problem around the already-stellar yet underexplored Zero-shot Image-based Retrieval performance of off-the-shelf models. Our fast-converging model keeps the single-domain performance while learning to extract similar representations from sketches. To this end we introduce our Semantic Anchors -- guiding embeddings learned from word-based semantic spaces and features from off-the-shelf models -- and combine them with our novel Anchored Contrastive Loss. Empirical evidence shows we can achieve state-of-the-art performance on all benchmark datasets while training for 100x less iterations than other methods.
\end{abstract}

\section{Introduction}
\label{sec:intro}

\begin{figure}[t]
    \centering
    \includegraphics[width=\linewidth]{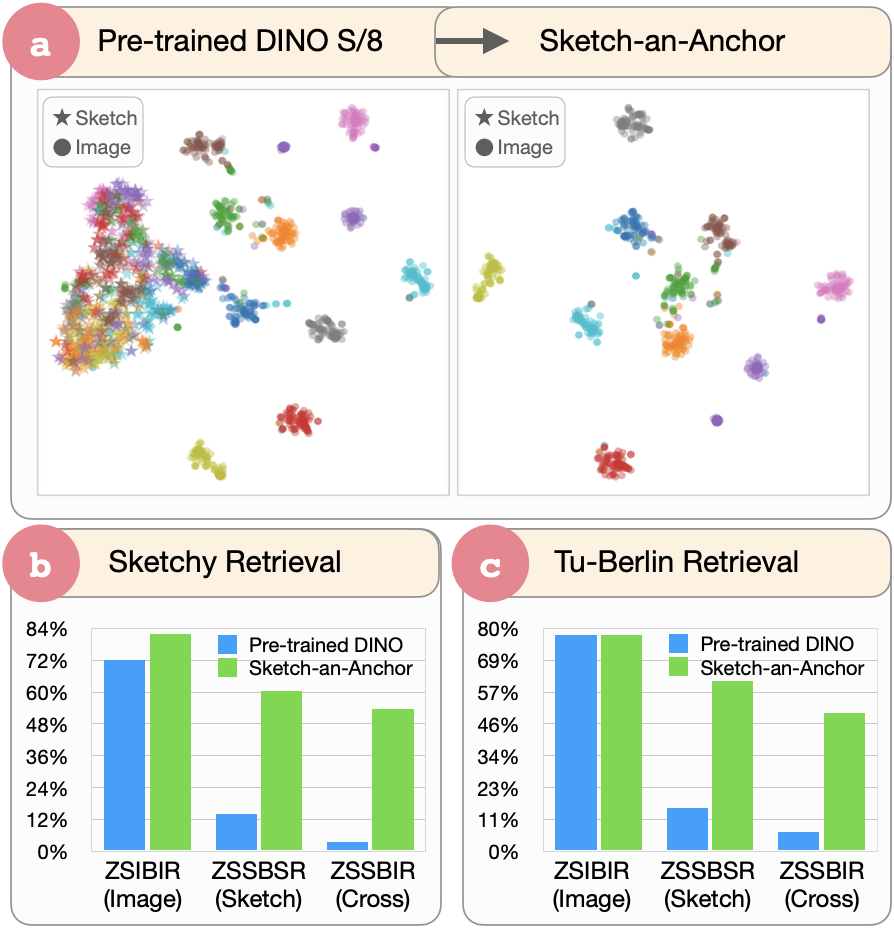}
    \caption{Our Sketch-an-Anchor model reframes the problem of Zero-shot Sketch-based Image Retrieval. Because pre-trained Self-Supervised models already perform well on Zero-shot Image-based Retrieval, our efforts should go into keeping this performance while mapping sketches to the same space. In (a) we compare our model and the pre-trained DINO S/8 that we start with by plotting a t-SNE projection of the feature vectors extracted from 10 classes of the TU-Berlin test set. In (b) and (c) we compare both models again, but show mAP (\%) on the Tu-Berlin and Sketchy Zero-shot test splits on three tasks, Image-based Image Retrieval, Sketch-based Sketch Retrieval and Sketch-based Image Retrieval.}
    \label{fig:teaser}
\end{figure}

As the world produces more photographic content and the internet takes the place of the real world in many fields, retrieval systems become ever more relevant. The task of Sketch-based Image Retrieval (SBIR) is a machine learning application that has been relevant for at least 20 years \cite{Ashley1995OldBlobPaperSBIR} and interest has only grown since; SBIR systems take sketched objects from users and ranks images on a database by semantic relevance.

While introducing Deep Learning methods to this task has led to great strides in performance, these systems also started depending on large datasets with both photos and sketches from the same category to perform well \cite{Dey2019Doodle2SearchZSSBIR, Shen2018HashingZSSBIR}. That motivated Zero-shot SBIR task (ZSSBIR) \cite{Shen2018HashingZSSBIR}; the protocol for ZSSBIR follows the common Zero-shot paradigm \cite{Xian2018ZeroShotSurvey}, the categories of images and sketches used for evaluation are not seen in training. This setup better mimics an open set scenario, which correspond to real applications for SBIR, where categories are vast and unpredictable: e.g. an e-commerce might add new products every day and an artist will search for different references for each creation.

Most studies break down the problem of ZSSBIR into two: domain alignment between images and sketches, inherited from SBIR, and generalization to unseen data, inherent to the zero-shot protocol. We argue that the latter of can be considerably simplified if we rely more strongly on pretrained models for photographs, following a trend that is gaining traction also in general Few-shot literature \cite{Hu2022Pushing}.

The field of Self-Supervised learning has also seen many proposals in recent years. The task of Self-supervision is to learn -- without labeled data -- general image representations that can be used for diverse downstream tasks; the approach often taken \cite{He2020MocoSelfS, Caron2021DINOSelfS, Chen2020Moco2SelfS, Grill2020BYOLSelfS, Chen2020SimCLRSelfS} is to train models effectively for image retrieval: semantically similar images should have similar representations.


We take DINO \cite{Caron2021DINOSelfS}, a recent self-supervised model and the best at k-NN classifier evaluation (where features are extracted from the frozen network to train a k-NN classifier for ImageNet), and test its Zero-shot Image-based Image Retrieval (ZSIBIR) performance without fine-tuning on the zero-shot image split of common ZSSBIR datasets. We've found that without any further training ZSIBIR performance is already excellent (See Fig. \ref{fig:teaser}), motivating us to rethink how we break down the ZSSBIR task. 

Domain alignment and generalization to unseen data are still at the core of a ZSSBIR solution, but, as we've discussed, the latter is solved or close to on the image domain by contrastive self-supervised models. With this information at hand, we've designed Sketch-an-Anchor, making it our goal to learn general sketch representations that match the existing image space of the pre-trained model.

In Fig. \ref{fig:teaser} (b,c) we graph a preliminary evaluation showing that Sketch-an-Anchor indeed succeeds at this goal. We've evaluated the DINO S/8 pre-trained model and our own finetuned one at within-domain and cross-domain retrieval, showing the former is already great at ZSIBIR (image domain) while ours maintains that performance at the same time as it significantly moves the needle on within-domain Sketch-based Sketch Retrieval and cross-domain ZSSBIR. The t-SNE projection in Fig. \ref{fig:teaser} (a) corroborates our finding, showing how the DINO space is discriminative for images but messy for sketches, and then how Sketch-an-Anchor keeps the image relationships while learning better representation for the sketches.

Rethinking the sub-tasks has another remarkable benefit: since we do not need to learn the ZSIBIR sub-task, our model requires only a few iterations, 99.1\% less iterations than the average ZSSBIR model to be exact, making it the fastest training ZSSBIR model in the literature.

To aid with our re-defined tasks we introduce the Semantic Anchors, a combination of the general Word2Vec \cite{Mikolov2013Word2Vec} semantic space and class centers based on the image representations output by the pre-trained model. We then use the Semantic Anchors and their pairwise similarities to design our novel Anchored Contrastive Loss, elevating a strong baseline model to SoTA performance. 


To summarize, our contributions are the following:
\begin{itemize}
    \addtolength\itemsep{-2mm}
    \item Re-think the ZSSBIR solution on matching sketch representations to an existing pre-trained space, since those are already good image retrieval models;
    \item The ability to train a model for ZSSBIR in under an epoch and achieve performance competitive with SoTA;
    \item Introduce Semantic Anchors, which learns from visual and word-based semantic spaces with a Graph Convolutional Net and guide the fast model adaptation;
    \item Our novel Anchored Contrastive Loss matches the sample similarity distribution to the anchors', further improving the task of matching sketch representations to the images'.
\end{itemize}

\section{Related Work}
\label{sec:related}

\noindent \textbf{Sketch-based Image Retrieval} has seen published solutions dating back to the 90s \cite{Ashley1995OldBlobPaperSBIR}, pre-deep-learning methods have worked with hand-crafted features and image edge-maps to extract comparable representations between images and sketches \cite{Eitz2011,Hu2013}, which were then combined with dictionary-learning methods such as Bag-of-Visual-Words. As they dominated computer vision, CNNs became the norm in SBIR too, with studies using multi-branch networks with contrastive \cite{wang2015sketch3d} or triplet losses \cite{qi2016, Bui2017compact, yu2016SketchMeShoe, sangkloy2016sketchy} and investigating the use of Edge Maps \cite{Bui2017compact} and whether to share weights between branches or not \cite{Bui2018sketching}.

\noindent \textbf{Zero-shot Sketch-based Image Retrieval.} The work of Shen \etal \cite{Shen2018HashingZSSBIR} was the first to introduce the task whilst also proposing a new hashing method to accelerate retrieval. Later works can be split by the side information they've used. The works of \cite{Dutta2020SEMPYCZSSBIR, Zhang2020SketchGCNZSSBIR, Dey2019Doodle2SearchZSSBIR, Chaudhuri2020SimplifiedZSSBIR, Deng2020PCSNZSSBIR} have used word-based semantics like we do to learn more general search spaces, while other models trust generative models to simulate unseen classes during training \cite{Dutta2021StyleGuideZSSBIR, Shen2018HashingZSSBIR, Yelamarthi2018FrameworkZSSBIR, Pandey2020StackedAEZSSBIR} or combine both \cite{Dutta2020SEMPYCZSSBIR}. The study of Liu \etal \cite{Liu2019SAKE}, SAKE, introduced the idea of using Knowledge Preservation to prevent pre-trained models from losing already acquired knowledge, and implemented their own custom distillation protocol to do so, many models follow similar protocols \cite{Wang2022PKSDZSSBIR,Tursun2022SBTKNet,Wang2021DSNZSSBIR,Tian2022TVTZSSBIR} or combinations of them with word-based semantic information. We share SAKE's authors' goal of avoiding knowledge loss but designed our approach to not need to rely on a computationally expensive teacher signal. We highlight three studies from the Knowledge Distillation set that are most similar to ours. TVT~\cite{Tian2022TVTZSSBIR} and PKSD~\cite{Wang2022PKSDZSSBIR} share our DINO backbone and the latter also explores the correlation between class centers, but their training protocol uses multiple ViT instances and they take 100x longer to converge. SBTKNet \cite{Tursun2022SBTKNet} is a recent study that makes use of class centers similarly to how we do our Semantic Anchors, which makes their model one of the fastest recent ones, training in under 25 epochs; however, 25 epochs is still more than 50 times longer than our Sketch-an-Anchor's training time. Our model is competitive against all of the previous studies while being simpler to implement, not requiring a teacher-student distillation setup or multiple network instances and, of course, being faster to train and to experiment with.

\section{The Sketch-an-Anchor Method}
\label{sec:method}

\begin{figure*}
    \centering
    \includegraphics[width=\linewidth]{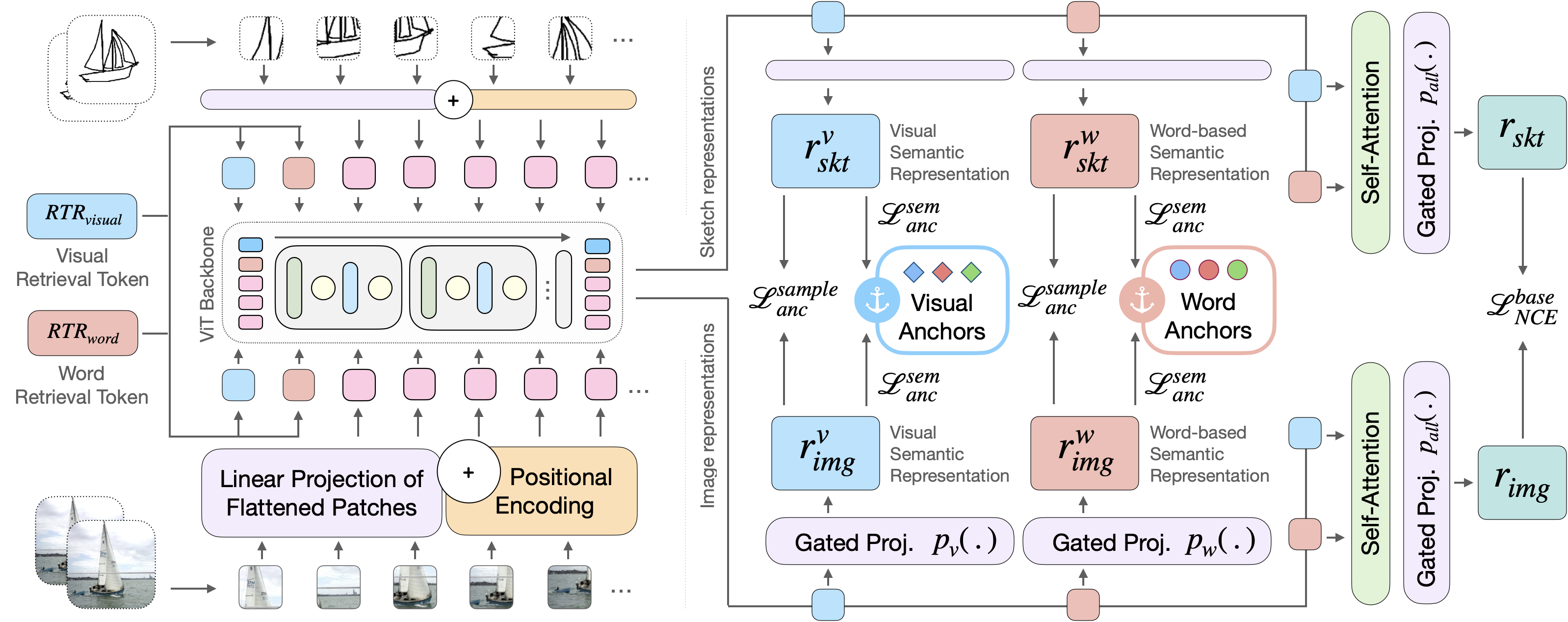}
    \caption{Diagram for the final Sketch-an-Anchor architecture. We start with a ViT S/8 backbone pre-trained with the DINO \cite{Caron2021DINOSelfS} protocol. To this backbone's input we include two learnable tokens, $RTR_{visual}$ and $RTR_{word}$. The outputs related to these tokens will make up the Visual and Word-based Representations, which are both trained to match images and sketches with $L_{anc}^{sample}$ and to match the semantic anchors with $L_{anc}^{sem}$. Finally, these representations are aggregated through self-attention and the final representation is trained to match domains with classic InfoNCE, $L_{NCE}^{base}$. The same network is used to separately extract features from sketches or photographic images.}
    \label{fig:model-arch}
\end{figure*}

In this section we describe the task and our solution. We take a constructivist explanation approach, building from backbone to final model. We start by introducing our baseline model, a strong and simple ZSSBIR solution, and then introduce and include the Semantic Anchors and Anchored Contrastive loss, additions that bring the model up to speed to compete with the best in the literature. A diagram for the complete architecture can be found in Fig. \ref{fig:model-arch}.

\subsection{Problem Setting}

Within Zero-shot Sketch-based Image Retrieval, a given model gets a sketched object as input and ranks images on a dataset based on how well they match the sketched object semantically. Under the zero-shot setting such model is evaluated with query sketches and retrieved images of objects that were not present in training. 

Formally, an SBIR dataset is split into two subsets to create the zero-shot setting, $\mathcal{D}_{\text{train}} = \{S_{\text{train}}, I_{\text{train}}, C_{\text{train}}\}$  and $\mathcal{D}_{\text{test}} = \{S_{\text{test}}, I_{\text{test}}, C_{\text{test}}\}$, where $S$ and $I$ are the set of labeled sketches and images for each subset, while $C$ are the classes of each subset. The zero-shot protocol is complete by ensuring there is no intersection between classes in each subset: $C_{\text{train}} \bigcap C_{\text{test}} = \emptyset$. 

At evaluation time, a ranked image is considered correct if it was labeled with the same class as the query sketch; this setup is commonly refereed to as Categorical SBIR and is the norm for ZSSBIR.

\subsection{A Strong Baseline}

As presented in Sec. \ref{sec:intro}, we are motivated by the already great performance of self supervised models on Zero-shot Image-based Retrieval (ZSIBR) and believe this performance can be transferred to ZSSBIR with little finetuning. We start with a Vision Transformer (ViT) backbone \cite{Dosovitskiy2021ViT}. A ViT encoder takes fixed-resolution ($R\times R$) patches of images as input, which are linearly projected to create a sequence of ``image tokens''; a learnable token $CLS$ is included in the sequence and all tokens are processed by $N$ sequential layers composed each of a pipeline of Layer Normalization, Multi-head Attention and feed-forward sublayers. Each layer is connected by skip-connections \cite{resnet}. 

These models are usually trained through supervised learning, but Caron \etal introduced DINO, a self-supervision protocol that makes use of multi-crop training \cite{Caron2020SwavSelfS}, a momentum encoder \cite{He2020MocoSelfS} and smaller ViT image patches to deliver representations that can make stellar k-NN classifiers. Because k-NN classifiers are essentially retrieval models we make a pre-trained DINO ViT our backbone model, specifically the compact DINO S/8~\cite{Caron2021DINOSelfS}. This is the same pre-trained backbone used by other SoTA ZSSBIR methods \cite{Wang2022PKSDZSSBIR, Tian2022TVTZSSBIR}, so it also makes comparison against those methods more fair. Other architectures/training protocols and the correlation of ZSIBR performance and our ZSSBIR results are in the Supp. Mat.. 

To adapt the pre-trained encoder $v(.)$ for the ZSSBIR task, the same model is used to extract representations from sketches $x_{skt}$ and images $x_{img}$. To compute this representation we include a new trainable token, $RTR$, together with the patch-based tokens from $x_{skt}$ or $x_{img}$ at the input. At the output, the processed $RTR$ goes through a projection head $p(.)$, implemented as gated linear projection \cite{Miech2017GatedProjection, Chen2021MultimodalGatedProjection}, yielding the representation vectors $r_{skt} = p(v(x_{skt}))$ and $r_{img} = p(v(x_{img}))$, for sketches and images. 

The inputs in $k$-th batch $B_k = \{S_{batch}, I_{batch}, C_{batch}\}$ are paired by class (at index $i$, $x_{skt}^i \in S_{batch}$ and $x_{img}^i \in I_{batch}$ are from the same class) and learning on the Baseline model is done through contrastive learning using InfoNCE:
\begin{equation}
    \mathcal{L}_{NCE}^{base}(B_k) = \dfrac{-1}{N}\sum^{N}_{i=0}log\dfrac{exp(r_{skt}^i\cdot r_{img}^i/\tau)}{\sum^{N}_{j=0}exp(r_{skt}^i\cdot r_{img}^j/\tau)}
\end{equation}
where $N$ is batch size, $\tau$ is the temperature hyper-parameter and controls the balance between a uniform-distributed embedding and a discriminative one \cite{Wang2020UnderstandingInfoNCE} and representations $r_{skt}$ and $r_{img}$ are compared with cosine similarity ($\cdot$). This loss then approximates sketches and images that share a label (numerator in the equation) and repels pairs that do not have the same label (denominator). 


We show in Sec. \ref{sec:experiments} how this Baseline is already a quickly adapting model for ZSSBIR which performs competitively.

\subsection{Semantic Anchors}

With our well-performing baseline at hand, we design additional components that lean into the strengths of the pre-trained model, starting with the \textit{Semantic Anchors}.

\paragraph{Word-based Embedding. }It is common practice in the ZSSBIR literature and other zero-shot tasks to include a generic semantic space as a guide to the model \cite{Dutta2020SEMPYCZSSBIR, Zhang2020SketchGCNZSSBIR, Deng2020PCSNZSSBIR}. That prevents training from adhering too close to the training classes, and thus learn more general concepts. The most common semantic embedding used is Word2Vec, which maps words into a semantic embedding space, and is trained by looking at word co-occurrence on millions of texts \cite{Mikolov2013Word2Vec}. We follow other methods and use it to map the words for each training class into this embedding space, creating what we call the \textit{Word Embedding}, $E_{word} = \{c^w_i\}$, with each $i$th vector matching the $i$th seen/training class.

\paragraph{Visual Embedding. } Because the chosen backbone has demonstrated retrieval performance for images, we argue that the space produced by its features is a good semantic embedding by itself, and it furthermore captures visual relationships that word embeddings cannot (See the Supp. Mat. for an investigation of the differences). We extract the representation for all images in the training sets and take the average vector of each class to make the \textit{Visual Embedding} $E_{visual} = \{c^v_i\}$. 

We collectively call these embeddings the \textit{Semantic Anchors} because they are responsible for grounding the output in both general semantic information -- the Word Embedding -- and in the semantic information already in the model -- the Visual Embedding. We explain how they are used in the next section.

\subsection{Anchored Contrastive Loss}

Given the Semantic Anchors in $E_{word}$ and $E_{visual}$, our goal is to make use of both spaces to tightly guide the finetuning process. This is done: (i) by matching hidden representations to their respective anchors and (ii) by matching hidden representations of sketches and images together while ``anchoring'' those to the ones in $E_{word}$ and $E_{visual}$. 

Both goals require architectural changes do the Baseline; to have two new hidden representations we swap the learnable token $RTR$ for two equivalents: $RTR_{word}$ and $RTR_{visual}$, responsible respectively for matching the Word and Visual embeddings at the output. These change yields two outputs from the $v(.)$ backbone, each of which are individually projected, producing $r_{skt}^w = p_w(v(x_{skt})_w)$ and $r_{skt}^v = p_v(v(x_{skt})_v)$ for each sketch and the equivalent pair $\{r_{img}^w, r_{img}^v\}$ for images, with the backbone $v(.)$ output subscript $\{w, v\}$ indicating word and visual, i.e. which of the $RTR$ tokens produced the representation.

To obtain the final single-vector retrieval representation, the pair $\{v(x_{skt})_w, v(x_{skt})_v\}$ (and the $x_{img}$ equivalent) is input into a Self-Attention layer $sa(.)$. We have used Self-Attention as designed by Sukhbaatar \etal \cite{Sukhbaatar2015End2EndSelfAttention} where the weights in $sa(.)$ learn to perform a dynamic weighted average of the vectors. We compute then $r_{skt} = sa(v(x_{skt})_w, v(x_{skt})_v)$ and $r_{img} = sa(v(x_{img})_w, v(x_{img})_v)$ for sketches and images. These aggregated vectors are then used to train the model as in the baseline with $\mathcal{L}^{base}_{NCE}$.

With a new architecture in place, to achieve goals (i, ii) we make use of the new pair of hidden representations $\{r_{d}^w, r_{d}^v\}$ ($d = \{skt, img\}$), combined with our novel Anchored Contrastive Loss, which will be presented now. From the Semantic Anchors $E_{word} = \{c^w_i\}$ and $E_{visual} = \{c^v_i\}$ we build similarity matrices for each embedding:
\begin{equation}
\begin{split}
    A^w & = \{a^w_{ij}\}, \text{with }a^w_{ij} = c^w_i.c^w_j \\
    A^v & = \{a^v_{ij}\}, \text{with }a^v_{ij} = c^v_i.c^v_j
\end{split}
\end{equation}

Then we use the similarities as probability distributions (by normalizing line-wise with softmax) and design Anchored Contrastive Loss as a cross-entropy function between the distribution of the inputs' similarities and the distribution of their anchors' similarities. This loss design, beyond being better aligned to smooth categorical retrieval when compared to InfoNCE, further preserves the geometrical properties of the Anchors while approximating similar and repelling dissimilar samples. We use this loss with two sets of inputs, with goal (i) in mind it is used to match the pair of hidden representations to their respective anchors:
\begin{equation}
    \dfrac{-1}{N}\sum^{N}_{i=0}\sum^{C}_{y=0}\sigma(a_{\gamma_i y})log \dfrac{exp(r_{d}^{s(i)} \cdot c^s_y / \tau)}{\sum_{j=0}^C exp(r_{d}^{s(i)} \cdot c^s_j / \tau)}
\end{equation}
where $\gamma_i$ represents the index of the anchor that matches the $i$-sh input in the batch and selects the anchor distance distribution for this sample; $\sigma$ is the softmax function; the subscripts $s = \{w, v\}$ and $d = \{skt, img\}$ are used to save space and show that the loss is applied in 4 variations, approximating sketches and images to the word and visual anchors. We refer to this loss as $\mathcal{L}_{anc}^{sem}$, the Anchored Contrastive loss that matches Semantic Anchors. 

We also make use of our new loss to achieve the second goal (ii) of matching sketches to images while ``grounded'' by the anchors:
\begin{equation}
    \dfrac{-1}{N}\sum^{N}_{i=0}\sum^{N}_{j=0}\sigma(a_{\gamma_i \gamma_j})log \dfrac{exp(r_{skt}^{s(i)} \cdot r_{img}^{s(j)} / \tau)}{\sum_{k=0}^N exp(r_{skt}^{s(i)} \cdot r_{img}^{s(k)} / \tau)}
\end{equation}
where $\gamma_j$ represents the index of the anchor that matches the $j$-sh input in the batch. Notice how, with our loss, sketches and images are matched pairwise and their hidden representations approximated or repelled based on the distribution of the anchors. We refer to this loss as $\mathcal{L}_{anc}^{sample}$, the Anchored Contrastive loss that matches sketches to images.

\subsection{Final Touches}

With the major components in place, we complete the model with two additions to the \textit{Semantic Anchors}; first, we make them more adaptable to the task, and second, we make the input embedding more dynamic.

\begin{figure}
    \centering
    \includegraphics[width=\linewidth]{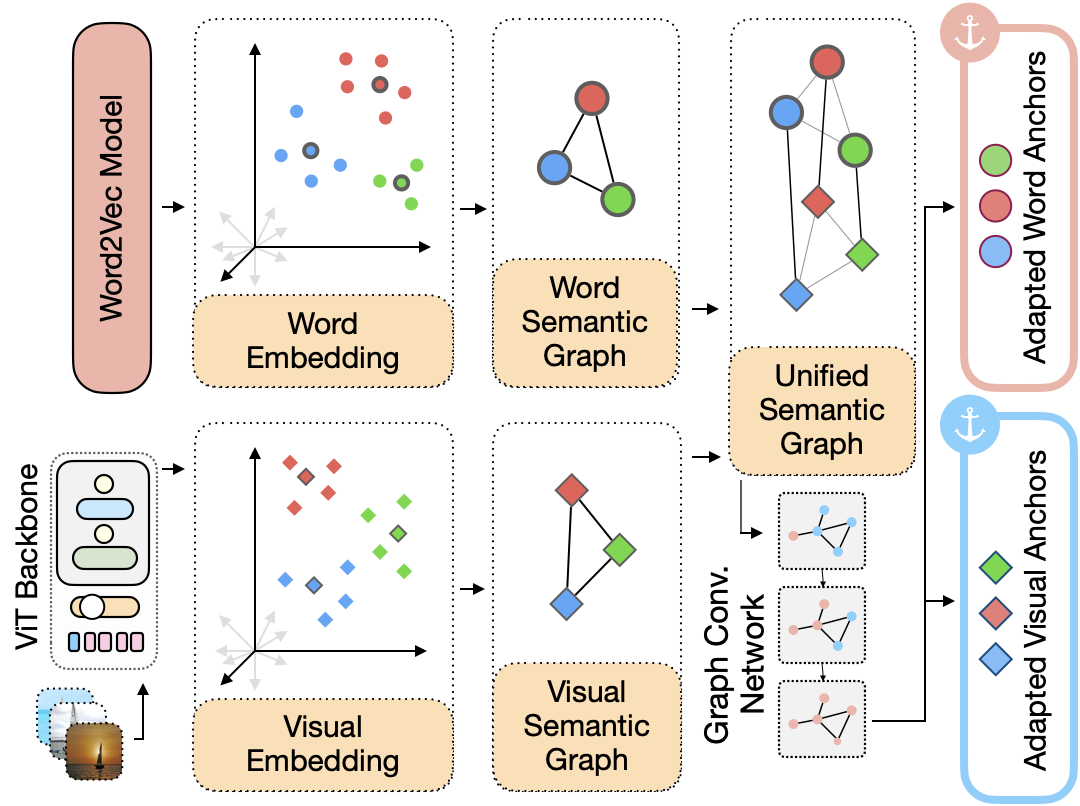}
    \caption{Diagram depicting how we learn the Adapted Semantic Anchors. We first create fully connected graphs with the original anchors and their similarity matrices, then we connect the two graphs by the matching categories. Finally we process the unified graph through a Graph Convolutional Network and input the Adapted Anchors to our model.}
    \label{fig:semantic-anchors}
\end{figure}

To let the anchors learn with the losses, we make use of a Graph Convolutional Network. First we create for each embedding a fully connected graph $\mathcal{G}^s = \{\mathcal{V}^s, \mathcal{E}^s\}$ composed of the embedding vectors as vertices $\mathcal{V}^s$ and the similarity matrix $A^s$ as the adjacency matrix that makes up the edges $\mathcal{E}^s$. These graphs, $\mathcal{G}^w$ and $\mathcal{G}^v$ for Word and Visual embedding, are unified by connecting matching classes together with edges with the same-weight as the self-connections (\eg the $c^w_i$ vector for ``cat'' is connected to the $c^v_i$ vector for ``cat''), this allows the embeddings to share knowledge. We normalize the edge values by applying softmax line-wise to the adjacency matrix and follow \cite{KipfWelling2016GCN} to build a GCN: \vspace{-0.1cm}
\begin{equation}
    (\mathcal{V}^{w+v})^{l+1} = \hat{A}^{w+v}(\mathcal{V}^{w+v})^{l}W^l
\end{equation}
where the superscript $w+v$ indicates these are the unified graphs, $\hat{A}^{w+v}$ is the normalized adjacency matrix, $l$ is a layer index, meaning these can be stacked, and $W^l$ are learnable weights for each layer. The output nodes $(\mathcal{V}^{w+v})^{L}$ are used as the \textit{Adapted Word} and \textit{Visual Anchors}; We've included a diagram in Fig. \ref{fig:semantic-anchors} depicting this process.

Finally, we make the input embeddings more dynamic and therefore more general. For each iteration, Word Embedding vectors are randomly swapped for one of the 10 closest word vectors in the full Word2Vec space; this means for example that the vector for ``cat'' can be swapped for the vectors for ``cats'' or ``feline''. For the Visual Embedding, instead of taking the class centers directly we use the center and standard deviation of each class to make a normal distribution and generate slight variants of class Anchors to use as input. 

The final Sketch-an-Anchor model is then trained with an equal weighted sum of the losses $\mathcal{L}^{base}_{NCE}$, the 4 variations of $\mathcal{L}_{anc}^{sem}$ and the 2 variations of $\mathcal{L}_{anc}^{samples}$. A full equation can be found of the Supp. Mat.

\section{Experiments}
\label{sec:experiments}

In this section we detail the benchmark datasets used, implementation details, as well as empirical evidence of our model's performance. We also show how we've used the baseline to determine how few iterations are necessary and demonstrate empirically our hypothesis that the task regards mapping sketches to the already excellent image space. 

\subsection{Datasets}

We follow most ZSSBIR literature and benchmark our model across three datasets, Sketchy \cite{sangkloy2016sketchy}, Tu-Berlin \cite{Eitz2012TuBerlin} and Quickdraw \cite{QD50}. Each dataset follows a similar benchmark protocol, being evaluated under categorical mean Average Precision (mAP) and Precision up to a rank $K$ ($P@K$). Features for test images are extracted beforehand to build a retrieval set and for each query sketch the cosine similarity between the sketch and the images' features is used to rank the images. 

\noindent \textbf{Sketchy} is composed of 75,471 sketches of 125 categories and originally had 12,500 natural images, but has since had this number expanded \cite{Shen2018HashingZSSBIR} 73,002 natural images. We consider two zero-shot splits, one mirroring SAKE \cite{Liu2019SAKE} where 25 classes were chosen randomly to compose the zero-shot test set; and another created by \cite{Yelamarthi2018FrameworkZSSBIR} with 21 zero-shot classes chosen specifically to avoid overlap with ImageNet; we refer to the latter as Non-Overlapping Sketchy (N.O. Sketchy).

\noindent \textbf{Tu-Berlin} contains 20k sketches from 250 categories and was extended \cite{Shen2018HashingZSSBIR} with 204k images matching such categories. An unrestricted selection of 30 zero-shot classes is used; for comparability we use the same selection as SAKE \cite{Liu2019SAKE}.

\noindent \textbf{QuickDraw} is the more recent and the largest set; In \cite{Dey2019Doodle2SearchZSSBIR} a selection of 110 of the 330 classes available was chosen based on how discriminative they were to humans; a set of 330k sketches and 204k images was built to compose QuickDraw Extended; we use the common 30 zero-shot classes.

\subsection{Implementation Details}

We use the Tensorflow 2 framework to build our models and run all of our experiments on a single NVIDIA RTX A5000; each experiment is repeated 5 times and we take the average metrics on evaluation. As mentioned, our backbone is a ViT-S/8, trained under the DINO protocol (DINO-S/8), this model comprises 12 layers with 6 attention heads in each. The gated linear projections used at the outputs all have 512 units. The GCN for the Semantic Anchors has 4 layers and each also has 512 units; we found the computational impact of adding the GCN to be negligible (a 0.5\% increase on iteration time on average). We multiply the gradient to the backbone by $0.1$ to accelerate learning of the projectors. 

All of our models are trained with a batch size of 16 for 1500 iterations. The Adam optimizer with learning rate $\eta=5\mathrm{e}{-6}$ is used and $\eta$ is warmed up (raised lineraly) to the base value within the first 150 iterations and is then decayed to $1\mathrm{e}{-6}$ with a cosine schedule. In the next Section (\ref{sec:iterations}) we explore our choice of number of iterations and learning rate.

\subsection{A Thousand Iterations are All You Need}
\label{sec:iterations}

One of our goals with this paper is to show that pre-trained models such as DINO can be quickly adapted to do ZSSBIR, so we designed a simple experiment to find how many iterations are needed to achieve good cross-domain performance. 

\begin{figure}
    \centering
    \includegraphics[width=\linewidth]{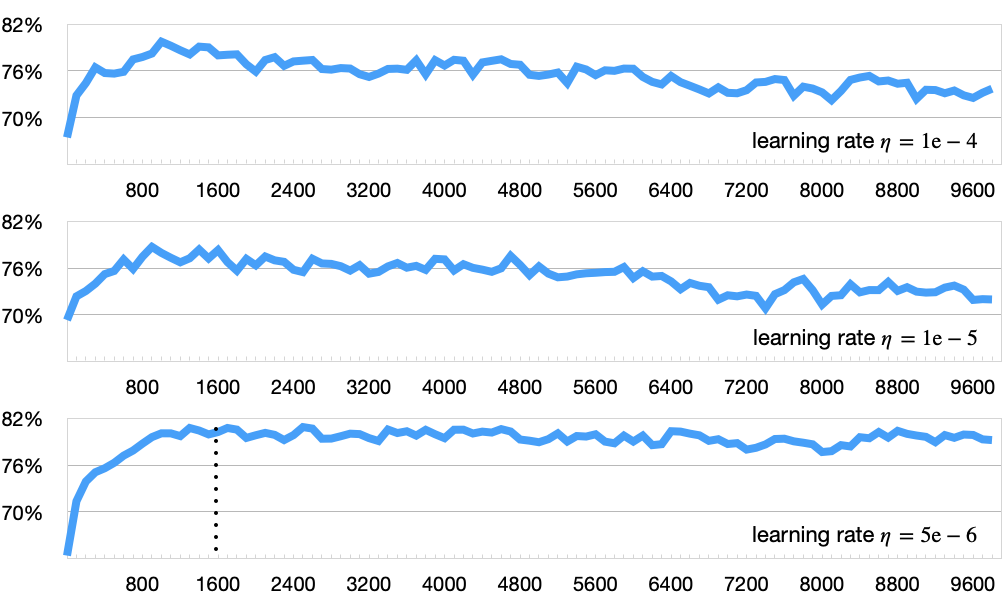}
    \caption{Evaluation on the Sketchy (Split \cite{Yelamarthi2018FrameworkZSSBIR}) validation set across 10 thousand iterations with the Baseline model. The x-axis shows iteration and the y-axis shows mAP (\%). We test three learning rate values and select to use $\eta=5\mathrm{e}{-6}$ and to stop training at 1500 iterations (dotted line on the bottom-most graph) for all our following experiments.}
    \label{fig:iteration-experiment}
\end{figure}

To this end a validation set for N.O. Sketchy, was created where 10 classes are separated from the training set, and from this set 64 images and sketches are randomly selected, making for a compact evaluation set. We then train our Baseline model with Adam and fixed learning rates for 10 thousand iterations. The results of this experiment can be found plotted in Fig. \ref{fig:iteration-experiment}. 

It is clear that regardless of learning rate, the models reach a peak mAP value fairly quickly after around one thousand iterations. We select the lower learning rate  $\eta=5\mathrm{e}{-6}$ for its stability and higher peak mAP, stopping our models at 1500 iterations instead of a thousand to allow for the inclusion of warm-up and a cosine schedule on the learning rate value. This low number of iterations corroborates our core hypothesis and our simple design.



\subsection{Comparison Against Other Methods}

\begin{table*}[h]
\centering
\resizebox{\textwidth}{!}{%
\begin{tabular}{lcccccccccccc}
\hline
\multirow{2}{*}{Methods} & \multicolumn{3}{l}{Sketchy} & \multicolumn{3}{l}{N.O. Sketchy} & \multicolumn{3}{l}{TU-Berlin} & \multicolumn{3}{l}{QuickDraw Extended} \\ 
 & \multicolumn{1}{l}{Ep.} & \multicolumn{1}{l}{mAP} & \multicolumn{1}{l}{P@100} & \multicolumn{1}{l}{Ep.} & \multicolumn{1}{l}{mAP@200} & \multicolumn{1}{l}{P@200} & \multicolumn{1}{l}{Ep.} & \multicolumn{1}{l}{mAP} & \multicolumn{1}{l}{P@100} & \multicolumn{1}{l}{Ep.} & \multicolumn{1}{l}{mAP} & \multicolumn{1}{l}{P@100} \\ \hline
SAKE \cite{Liu2019SAKE} & 20 & 54.7 & 69.2 & 20 & 49.7 & 59.8 & 20 & 47.5 & 59.9 & - & - & - \\
DSN \cite{Wang2021DSNZSSBIR} & 10 & 58.3 & 70.4 & - & - & - & 10 & 48.4 & 58.6 & - & - & - \\
PCSN \cite{Deng2020PCSNZSSBIR} & 20 & 52.3 & 61.6 & - & - & - & 20 & 42.4 & 51.7 & - & - & - \\
SPCYC \cite{Dutta2020SEMPYCZSSBIR} & - & 34.9 & 46.3 & - & - & - & - & 29.7 & 42.6 & - & \textbf{17.7} & 25.5 \\
SGCN \cite{Zhang2020SketchGCNZSSBIR} & - & - & - & - & \textbf{56.8} & 48.7 & - & 32.4 & 50.5 & - & - & - \\
SBTK \cite{Tursun2022SBTKNet} & 25 & 55.3 & 69.8 & 25 & 50.2 & 59.6 & 25 & 48.0 & {\ul 60.8} & 25 & 11.9 & - \\
TVT \cite{Tian2022TVTZSSBIR} & 50 & 64.8 & \textbf{79.6} & 50 & 53.1 & 61.8 & 50 & 48.4 & \textbf{66.2} & 50 & 14.9 & \textbf{29.9} \\
BDA \cite{Chaudhuri2022BDASketRetZSSBIR} & 50 & 43.7 & 51.4 & 50 & 55.6 & 45.8 & 50 & 37.4 & 50.4 & 50 & {\ul 15.4} & 28.6 \\
PSKD \cite{Wang2022PKSDZSSBIR} & 20 & \textbf{68.8} & {\ul 78.6} & 20 & {\ul 56.0} & \textbf{64.5} & 20 & \textbf{50.2} & \textbf{66.2} & 20 & 15.0 & {\ul 29.7} \\
\textbf{Ours} & \textbf{0.4} & {\ul 67.1} & 76.2 & \textbf{0.4} & 53.5 & {\ul 63.0} & \textbf{0.1} & {\ul 49.5} & {\ul 60.8} & \textbf{0.1} & 14.8 & 21.2 \\ \hline
\end{tabular}%
}
\caption{Comparison of Sketch-an-Anchor against nine methods from the literature at the ZSSBIR task. We compare on both splits of the Sketchy, on Tu-Berlin and on Quickdraw Extended. Missing results were not reported in the original papers. We also highlight the number of Epochs (Ep.) each model takes to train when reported and bold and underline the best and second-best results for all metrics.}
\label{tab:main-results}
\end{table*}

We compare our Sketch-an-Anchor model against nine of the best methods in the literature. We furthermore complement the comparison with the Generalized ZSSBIR sub-task. 

\begin{figure}
    \centering
    \includegraphics[width=\linewidth]{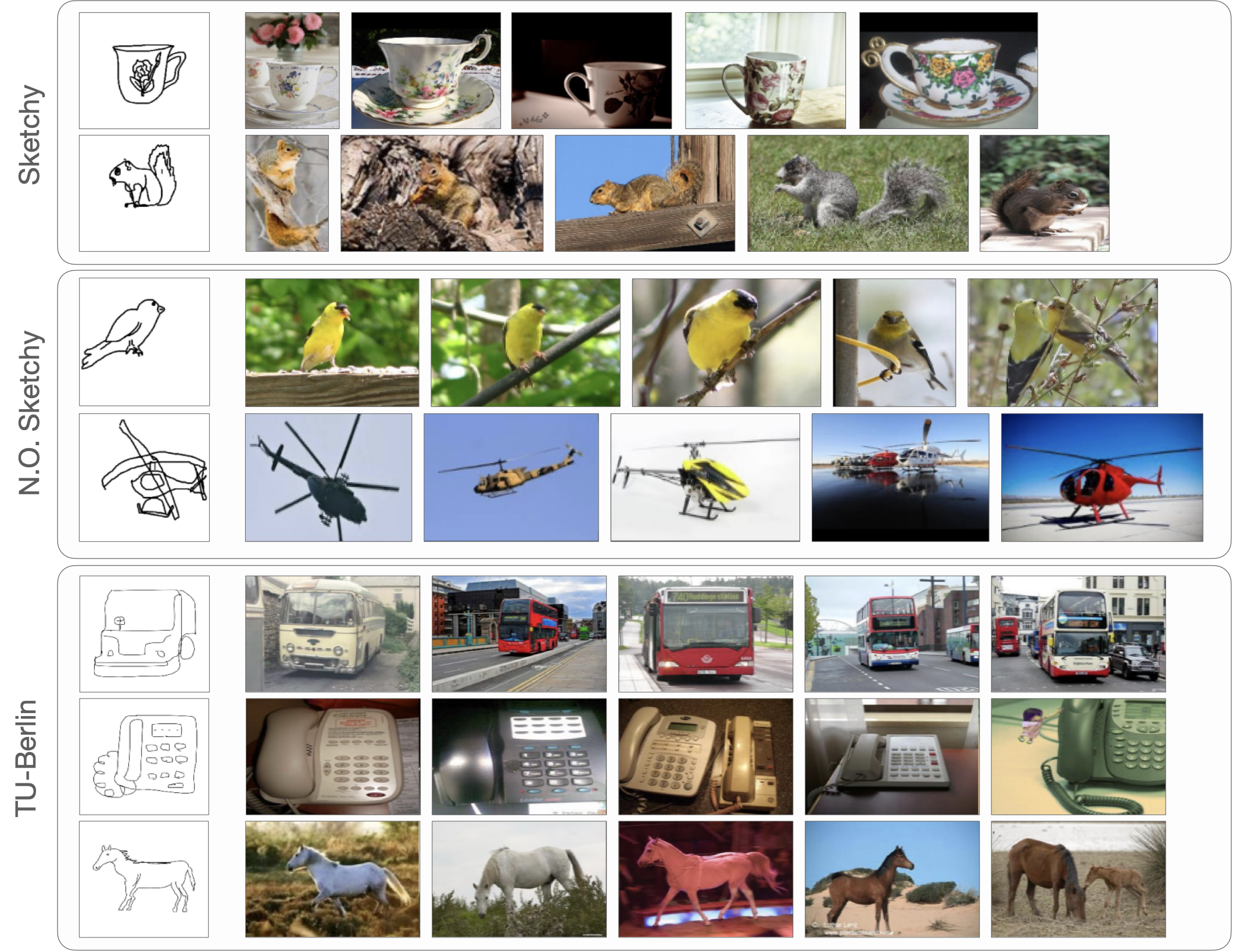}
    \caption{Visual ZSSBIR results. For each query sketch we show the 5 top images retrieved. The first two rows are from the Sketchy dataset's original split, the next two from the N.O. split \cite{Yelamarthi2018FrameworkZSSBIR} and the last three from Tu-Berlin.}
    \label{fig:sbir}
\end{figure}

The results for classic ZSSBIR can be found in Table \ref{tab:main-results}, and some qualitative ZSSBIR results for our model are on Fig. \ref{fig:sbir}. We can see that our results on ZSSBIR are very competitive against the SoTA, often being around 1\% shy of the best performing model in each metric. We've included the number of training epochs (Ep.) for methods that report it and highlight that our method is trained end-to-end in 1500 iterations, being on average 110 times faster to train (0.9\% the number of iterations of other methods). 

We take special note of the comparison against TVT \cite{Tian2022TVTZSSBIR} and PSKD \cite{Wang2022PKSDZSSBIR} since both share a backbone (DINO S/8) with our model but do not enjoy the benefit of its fast adaptation properties. TVT implements three ViT instances and trains for 50 epochs while PSKD uses two and trains for 20 epochs; they both use knowledge distillation to avoid losing the knowledge already in the pre-trained model. Our model doesn't need distillation as the protocol of few iterations and low learning rate by itself keeps the knowledge from deprecation, avoiding the ``catastrophic forgetting'' phenomenon. Given the shared backbone, we go beyond comparing epochs and directly compare the number of parameters and computational cost of one training (and one inference) iteration in Table \ref{tab:flops}. Here we can see that beyond needing 50-80 times less iterations, each training iteration in Sketch-an-Anchor uses half the computational resources of these competitors.

\begin{table}[]
\centering
\begin{tabular}{lccc}
Methods & Param. & Train. FLOPS & Inf. FLOPS \\ \hline
TVT \cite{Tian2022TVTZSSBIR} & 227 M & 1.4 T & 22.7 G \\
PSKD \cite{Wang2022PKSDZSSBIR} & 127 M & 1.4 T & 22.7 G \\
Ours & \textbf{26 M} & \textbf{0.7 T} & \textbf{22.7 G} \\ \hline
\end{tabular}
\caption{Computational and memory cost comparison of Sketch-an-Anchor against the two competitors that also use the DINO S/8 backbone. Training FLOPS considers a single iteration with batch size 16. Inference FLOPS considers the encoding one sample.}
\label{tab:flops}
\end{table}



The task of Generalized ZSSBIR, GZSSBIR, considers the search corpus to have images from both seen and unseen classes and we follow the literature to take 20\% of the images from each class in the training set and adding it to the test set instead; queries remain being restricted to unseen images. Not all methods test under this task, we've included our results against three of the nine aforementioned competitors in Table \ref{tab:generalized}. The results here reflect the ones in classic ZSSBIR, with our model producing competitive results while being fast to train.

\begin{table}[]
\centering
\begin{tabular}{lcccc}
\multirow{2}{*}{Methods} & \multicolumn{2}{l}{Sketchy} & \multicolumn{2}{l}{TU-Berlin} \\
 & \multicolumn{1}{l}{mAP} & \multicolumn{1}{l}{P@100} & \multicolumn{1}{l}{mAP} & \multicolumn{1}{l}{P@100} \\ \hline
SEM-PCYC \cite{Dutta2020SEMPYCZSSBIR} & 30.7 & 29.6 & 19.2 & 21.8 \\
SBTKNet \cite{Tursun2022SBTKNet} & {\ul 51.5} & {\ul 57.2} & \textbf{33.4} & \textbf{49.4} \\
BDA \cite{Chaudhuri2022BDASketRetZSSBIR} & 33.8 & 41.3 & 25.1 & 35.7 \\
Ours & \textbf{52.5} & \textbf{59.0} & {\ul 29.0} & {\ul 38.1} \\ \hline
\end{tabular}
\caption{Comparison of Sketch-an-Anchor against 3 methods from the literature at the Generalized ZSSBIR task. We compare on the random split of Sketchy and on Tu-Berlin to match existing reports.}
\label{tab:generalized}
\end{table}

\subsection{An Image is Worth an Entire Class}

We argue that when using powerful pre-trained models ZSSBIR can be taken as the task of mapping sketches to the image space while keeping representations general enough for the zero-shot protocol. Our Semantic Anchors, Visual Anchors specially, follow this thread, we introduce guiding points in the space so that images can stay within their representations and sketches can follow.

To further test this hypothesis we ask: if the sketches are the ones that need to be mapped, could we use only a representative sub-set of the images to train the model? Furthermore, how representative the Visual Anchors (class centers) are of the image space?

\begin{figure*}
    \centering
    \includegraphics[width=\linewidth]{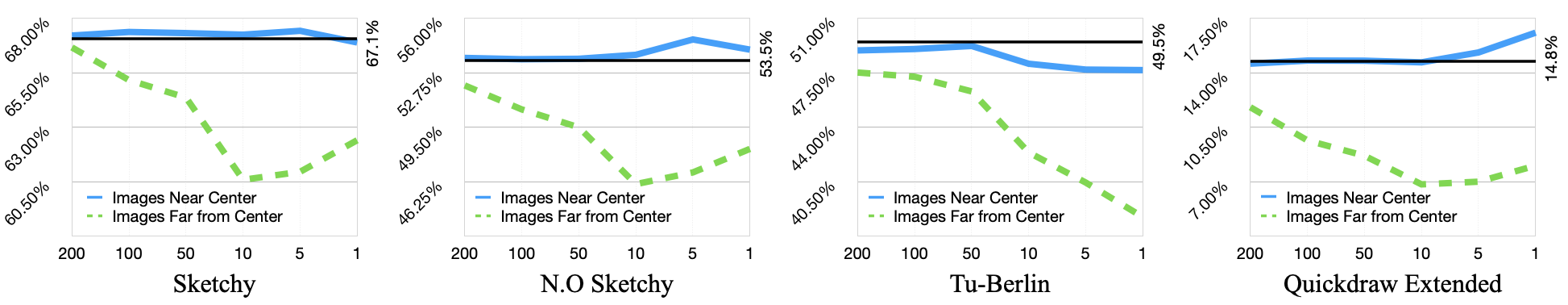}
    \caption{Experiment with Image Selection. For each model we select the $N$ (x-axis) images closer (blue, filled line) to their corresponding Visual Anchor or the $N$ images further from them (green, dotted line). For each selection we train Sketch-an-Anchor using the full sketch set and the selected images. The y-axis shows mAP for each model (mAP@200 for N.O. Sketchy). Black line shows original performance.}
    \label{fig:filter-experiment}
\end{figure*}

To answer these questions we've designed a new experiment: instead of taking the entire image set to train our model, we make two ``opposing'' sets; in one we select only the $N$ images from each class closer to their Visual Anchors, in the other we select the $N$ images farther from them. We repeat this with $N=\{200, 100, 50, 10, 5, 1\}$ and show the results in Fig. \ref{fig:filter-experiment}. 

As is clear from the results, Sketch-an-Anchor retains its competitive performance even when only training with a single image from each class as long as this image matches its Visual Anchor. In both Sketchy splits performance is almost unchanged across all number of images and while the Tu-Berlin set needs at least 50 images to retain performance, that is still less than 10\% of the original set. Finally, using only a single image improves the result on Quickdraw Extended to 16.6\%, which we believe is due to how noisy the full image set is on this set.

The fact that the images closer to the Visual Anchors are able to train the model to the same performance level shows how relevant this side information is to the ZSSBIR task and our fast adapting Sketch-an-Anchor.

\subsection{Ablation Study}

We follow the description in Sec. \ref{sec:method} and build our ablation studies in the same fashion as we've built the model, starting from the Baseline, including the Semantic Anchors and Anchored Contrastive loss (A) and finally the GCN and Randomized Semantic Input (B).

\begin{table}[]
\centering
\resizebox{\columnwidth}{!}{%
\begin{tabular}{lccccccc}
ID & \multicolumn{1}{l}{$E_v$} & \multicolumn{1}{l}{$E_w$} & \multicolumn{1}{l}{$\mathcal{L}_{anc}$} & \multicolumn{1}{l}{R.} & \multicolumn{1}{l}{GCN} & \multicolumn{1}{l}{mAP} & \multicolumn{1}{l}{P@100} \\ \hline
Base & \textbf{} & \textbf{} & \textbf{} & \textbf{} & \textbf{} & 47.06 & 59.65 \\
A1 & \checkmark &  & \checkmark  &  &  & 47.74 & 59.78 \\
A2 & \checkmark & \checkmark & \checkmark &  &  & 48.10 & 59.59 \\
B1 & \checkmark & \checkmark & \checkmark & \checkmark &  & 48.66 & 60.03 \\
B2 & \checkmark & \checkmark & \checkmark &  & \checkmark & 48.96 & 60.22 \\
\textbf{Ours} & \checkmark & \checkmark & \checkmark & \checkmark & \checkmark & \textbf{49.50} & \textbf{60.82}
\end{tabular}%
}
\caption{Evaluation of ablated models on the Tu-Berlin dataset, reported with Mean Average Precision (mAP) and Precision at 100. We refer to the Semantic Anchors as $E_v$ and $E_w$, and refer to Randomized Semantic Input as R.. 
}
\label{tab:ablations}
\end{table}

The results for each of our ablated models are depicted in Table \ref{tab:ablations}. Experiments were conducted with the harder Tu-Berlin dataset. We can see from the studies that adding each of the Semantic Anchors with our new loss brings improvements to the baseline and completing the model, Randomized Semantic Input (R.) and the GCN both add individual improvements and work best when used together; this is an effect of how randomization contributes to a GCN that produces more general Adapted Semantic Anchors. All together we add 2.9\% to the baseline's mAP, enough to make it the second-best Tu-Berlin mAP in the literature.

\section{Conclusion}

We've presented Sketch-an-Anchor, a new competitive model for ZSSBIR. We started with the insight that the problem of Zero-shot Image Retrieval is almost solved by Self-Supervision models such as DINO; with this in mind, we can stop training models from scratch or for long by focusing on mapping sketches to the already excellent image space. We've shown how to train a baseline model that is already among the best at ZSSBIR and introduced our Semantic Anchors and the Anchored Contrastive Loss to push this model to be competitive with SoTA performance. Also, our experiments have shown that our model only needs a few (or even only one) images per class to match sketches to the image space and that each of our additions to the baseline brings improvements in retrieval performance. Such findings may significantly change the way future approaches tackle the problem, allowing for experimentation with new models at a much higher rate. Future work should explore if our findings apply to the Domain Generalization and Few-shot learning fields of study, both of which share some of the challenges we've addressed on ZSSBIR.

{\small
\bibliographystyle{ieee_fullname}
\bibliography{arxiv_paper}

\begin{thebibliography}{10}\itemsep=-1pt

\bibitem{Ashley1995OldBlobPaperSBIR}
Jonathan Ashley, Myron Flickner, James Hafner, Denis Lee, Wayne Niblack, and
  Dragutin Petkovic.
\newblock The query by image content (qbic) system.
\newblock In {\em Proceedings of the 1995 ACM SIGMOD International conference
  on Management of data}, page 475, 1995.

\bibitem{Bui2017compact}
Tu Bui, Leo Ribeiro, Moacir Ponti, and John Collomosse.
\newblock Compact descriptors for sketch-based image retrieval using a triplet
  loss convolutional neural network.
\newblock {\em CVIU}, 2017.

\bibitem{Bui2018sketching}
Tu Bui, Leo Ribeiro, Moacir Ponti, and John Collomosse.
\newblock Sketching out the details: Sketch-based image retrieval using
  convolutional neural networks with multi-stage regression.
\newblock {\em Computers \& Graphics}, 71:77--87, 2018.

\bibitem{Caron2020SwavSelfS}
Mathilde Caron, Ishan Misra, Julien Mairal, Priya Goyal, Piotr Bojanowski, and
  Armand Joulin.
\newblock Unsupervised learning of visual features by contrasting cluster
  assignments.
\newblock {\em Advances in Neural Information Processing Systems},
  33:9912--9924, 2020.

\bibitem{Caron2021DINOSelfS}
Mathilde Caron, Hugo Touvron, Ishan Misra, Herv{\'e} J{\'e}gou, Julien Mairal,
  Piotr Bojanowski, and Armand Joulin.
\newblock Emerging properties in self-supervised vision transformers.
\newblock In {\em Proceedings of the IEEE/CVF International Conference on
  Computer Vision}, pages 9650--9660, 2021.

\bibitem{Chaudhuri2020SimplifiedZSSBIR}
Ushasi Chaudhuri, Biplab Banerjee, Avik Bhattacharya, and Mihai Datcu.
\newblock A simplified framework for zero-shot cross-modal sketch data
  retrieval.
\newblock In {\em Proceedings of the IEEE/CVF Conference on Computer Vision and
  Pattern Recognition Workshops}, pages 182--183, 2020.

\bibitem{Chaudhuri2022BDASketRetZSSBIR}
Ushasi Chaudhuri, Ruchika Chavan, Biplab Banerjee, Anjan Dutta, and Zeynep
  Akata.
\newblock Bda-sketret: Bi-level domain adaptation for zero-shot sbir.
\newblock {\em arXiv preprint arXiv:2201.06570}, 2022.

\bibitem{Chen2021MultimodalGatedProjection}
Brian Chen, Andrew Rouditchenko, Kevin Duarte, Hilde Kuehne, Samuel Thomas,
  Angie Boggust, Rameswar Panda, Brian Kingsbury, Rogerio Feris, David Harwath,
  et~al.
\newblock Multimodal clustering networks for self-supervised learning from
  unlabeled videos.
\newblock In {\em Proceedings of the IEEE/CVF International Conference on
  Computer Vision}, pages 8012--8021, 2021.

\bibitem{Chen2020SimCLRSelfS}
Ting Chen, Simon Kornblith, Mohammad Norouzi, and Geoffrey Hinton.
\newblock A simple framework for contrastive learning of visual
  representations.
\newblock In {\em International conference on machine learning}, pages
  1597--1607. PMLR, 2020.

\bibitem{Chen2020Moco2SelfS}
Xinlei Chen, Haoqi Fan, Ross Girshick, and Kaiming He.
\newblock Improved baselines with momentum contrastive learning.
\newblock {\em arXiv preprint arXiv:2003.04297}, 2020.

\bibitem{Deng2020PCSNZSSBIR}
Cheng Deng, Xinxun Xu, Hao Wang, Muli Yang, and Dacheng Tao.
\newblock Progressive cross-modal semantic network for zero-shot sketch-based
  image retrieval.
\newblock {\em IEEE Transactions on Image Processing}, 29:8892--8902, 2020.

\bibitem{Dey2019Doodle2SearchZSSBIR}
Sounak Dey, Pau Riba, Anjan Dutta, Josep Llados, and Yi-Zhe Song.
\newblock Doodle to search: Practical zero-shot sketch-based image retrieval.
\newblock In {\em Proceedings of the IEEE/CVF Conference on Computer Vision and
  Pattern Recognition}, pages 2179--2188, 2019.

\bibitem{Dosovitskiy2021ViT}
Alexey Dosovitskiy, Lucas Beyer, Alexander Kolesnikov, Dirk Weissenborn,
  Xiaohua Zhai, Thomas Unterthiner, Mostafa Dehghani, Matthias Minderer, Georg
  Heigold, Sylvain Gelly, Jakob Uszkoreit, and Neil Houlsby.
\newblock An image is worth 16x16 words: Transformers for image recognition at
  scale.
\newblock In {\em International Conference on Learning Representations}, 2021.

\bibitem{Dutta2020SEMPYCZSSBIR}
Anjan Dutta and Zeynep Akata.
\newblock Semantically tied paired cycle consistency for any-shot sketch-based
  image retrieval.
\newblock {\em Int. J. Comput. Vis.}, 128(10):2684--2703, 2020.

\bibitem{Dutta2021StyleGuideZSSBIR}
Titir Dutta, Anurag Singh, and Soma Biswas.
\newblock Styleguide: Zero-shot sketch-based image retrieval using style-guided
  image generation.
\newblock {\em {IEEE} Trans. Multim.}, 23:2833--2842, 2021.

\bibitem{Eitz2012TuBerlin}
Mathias Eitz, James Hays, and Marc Alexa.
\newblock How do humans sketch objects?
\newblock {\em Proc. ACM SIGGRAPH}, pages 44:1--44:10, 2012.

\bibitem{Eitz2011}
Mathias Eitz, Kristian Hildebrand, Tamy Boubekeur, and Marc Alexa.
\newblock Sketch-based image retrieval: Benchmark and bag-of-features
  descriptors.
\newblock {\em IEEE Trans. Visualization and Computer Graphics},
  17(11):1624--1636, 2011.

\bibitem{Grill2020BYOLSelfS}
Jean-Bastien Grill, Florian Strub, Florent Altch{\'e}, Corentin Tallec, Pierre
  Richemond, Elena Buchatskaya, Carl Doersch, Bernardo Avila~Pires, Zhaohan
  Guo, Mohammad Gheshlaghi~Azar, et~al.
\newblock Bootstrap your own latent-a new approach to self-supervised learning.
\newblock {\em Advances in neural information processing systems},
  33:21271--21284, 2020.

\bibitem{He2020MocoSelfS}
Kaiming He, Haoqi Fan, Yuxin Wu, Saining Xie, and Ross Girshick.
\newblock Momentum contrast for unsupervised visual representation learning.
\newblock In {\em Proceedings of the IEEE/CVF conference on computer vision and
  pattern recognition}, pages 9729--9738, 2020.

\bibitem{resnet}
Kaiming He, Xiangyu Zhang, Shaoqing Ren, and Jian Sun.
\newblock Deep residual learning for image recognition.
\newblock In {\em Proc. CVPR}, pages 770--778, 2016.

\bibitem{Hu2013}
Rui Hu and John Collomosse.
\newblock A performance evaluation of gradient field hog descriptor for sketch
  based image retrieval.
\newblock {\em CVIU}, 117(7):790--806, 2013.

\bibitem{Hu2022Pushing}
Shell~Xu Hu, Da Li, Jan St{\"u}hmer, Minyoung Kim, and Timothy~M Hospedales.
\newblock Pushing the limits of simple pipelines for few-shot learning:
  External data and fine-tuning make a difference.
\newblock In {\em Proceedings of the IEEE/CVF Conference on Computer Vision and
  Pattern Recognition}, pages 9068--9077, 2022.

\bibitem{KipfWelling2016GCN}
Thomas~N Kipf and Max Welling.
\newblock Semi-supervised classification with graph convolutional networks.
\newblock {\em arXiv preprint arXiv:1609.02907}, 2016.

\bibitem{Liu2019SAKE}
Qing Liu, Lingxi Xie, Huiyu Wang, and Alan~L Yuille.
\newblock Semantic-aware knowledge preservation for zero-shot sketch-based
  image retrieval.
\newblock In {\em Proceedings of the IEEE/CVF International Conference on
  Computer Vision}, pages 3662--3671, 2019.

\bibitem{Miech2017GatedProjection}
Antoine Miech, Ivan Laptev, and Josef Sivic.
\newblock Learnable pooling with context gating for video classification.
\newblock {\em arXiv preprint arXiv:1706.06905}, 2017.

\bibitem{Mikolov2013Word2Vec}
Tomas Mikolov, Ilya Sutskever, Kai Chen, Greg~S Corrado, and Jeff Dean.
\newblock Distributed representations of words and phrases and their
  compositionality.
\newblock {\em Advances in neural information processing systems}, 26, 2013.

\bibitem{Pandey2020StackedAEZSSBIR}
Anubha Pandey, Ashish Mishra, Vinay~Kumar Verma, Anurag Mittal, and Hema~A.
  Murthy.
\newblock Stacked adversarial network for zero-shot sketch based image
  retrieval.
\newblock In {\em {IEEE} Winter Conference on Applications of Computer Vision,
  {WACV} 2020, Snowmass Village, CO, USA, March 1-5, 2020}, pages 2529--2538.
  {IEEE}, 2020.

\bibitem{qi2016}
Yonggang Qi, Yi-Zhe Song, Honggang Zhang, and Jun Liu.
\newblock Sketch-based image retrieval via siamese convolutional neural
  network.
\newblock In {\em Proc. ICIP}, pages 2460--2464. IEEE, 2016.

\bibitem{QD50}
The {Q}uick, {D}raw! {D}ataset.
\newblock \url{https://github.com/googlecreativelab/quickdraw-dataset}, 2018.
\newblock Accessed: 2018-10-11.

\bibitem{sangkloy2016sketchy}
Patsorn Sangkloy, Nathan Burnell, Cusuh Ham, and James Hays.
\newblock The sketchy database: learning to retrieve badly drawn bunnies.
\newblock {\em ACM Trans. Graphics (TOG)}, 35(4):119, 2016.

\bibitem{Shen2018HashingZSSBIR}
Yuming Shen, Li Liu, Fumin Shen, and Ling Shao.
\newblock Zero-shot sketch-image hashing.
\newblock In {\em 2018 {IEEE} Conference on Computer Vision and Pattern
  Recognition, {CVPR} 2018, Salt Lake City, UT, USA, June 18-22, 2018}, pages
  3598--3607. Computer Vision Foundation / {IEEE} Computer Society, 2018.

\bibitem{Sukhbaatar2015End2EndSelfAttention}
Sainbayar Sukhbaatar, Arthur Szlam, Jason Weston, and Rob Fergus.
\newblock End-to-end memory networks.
\newblock In Corinna Cortes, Neil~D. Lawrence, Daniel~D. Lee, Masashi Sugiyama,
  and Roman Garnett, editors, {\em Advances in Neural Information Processing
  Systems 28: Annual Conference on Neural Information Processing Systems 2015,
  December 7-12, 2015, Montreal, Quebec, Canada}, pages 2440--2448, 2015.

\bibitem{Tian2022TVTZSSBIR}
Jialin Tian, Xing Xu, Fumin Shen, Yang Yang, and Heng~Tao Shen.
\newblock Tvt: Three-way vision transformer through multi-modal hypersphere
  learning for zero-shot sketch-based image retrieval.
\newblock {\em Proceedings of the AAAI Conference on Artificial Intelligence},
  36(2):2370--2378, 2022.

\bibitem{Tursun2022SBTKNet}
Osman Tursun, Simon Denman, Sridha Sridharan, Ethan Goan, and Clinton Fookes.
\newblock An efficient framework for zero-shot sketch-based image retrieval.
\newblock {\em Pattern Recognition}, 126:108528, 2022.

\bibitem{wang2015sketch3d}
Fang Wang, Le Kang, and Yi Li.
\newblock Sketch-based 3d shape retrieval using convolutional neural networks.
\newblock In {\em Proc. CVPR}, pages 1875--1883. IEEE, 2015.

\bibitem{Wang2022PKSDZSSBIR}
Kai Wang, Yifan Wang, Xing Xu, Xin Liu, Weihua Ou, and Huimin Lu.
\newblock Prototype-based selective knowledge distillation for zero-shot sketch
  based image retrieval.
\newblock In {\em Proceedings of the 30th ACM International Conference on
  Multimedia}, pages 601--609, 2022.

\bibitem{Wang2020UnderstandingInfoNCE}
Tongzhou Wang and Phillip Isola.
\newblock Understanding contrastive representation learning through alignment
  and uniformity on the hypersphere.
\newblock In {\em International Conference on Machine Learning}, pages
  9929--9939. PMLR, 2020.

\bibitem{Wang2021DSNZSSBIR}
Zhipeng Wang, Hao Wang, Jiexi Yan, Aming Wu, and Cheng Deng.
\newblock Domain-smoothing network for zero-shot sketch-based image retrieval.
\newblock In Zhi-Hua Zhou, editor, {\em Proceedings of the Thirtieth
  International Joint Conference on Artificial Intelligence, {IJCAI-21}}, pages
  1143--1149. International Joint Conferences on Artificial Intelligence
  Organization, 8 2021.
\newblock Main Track.

\bibitem{Xian2018ZeroShotSurvey}
Yongqin Xian, Christoph~H Lampert, Bernt Schiele, and Zeynep Akata.
\newblock Zero-shot learning—a comprehensive evaluation of the good, the bad
  and the ugly.
\newblock {\em IEEE transactions on pattern analysis and machine intelligence},
  41(9):2251--2265, 2018.

\bibitem{Yelamarthi2018FrameworkZSSBIR}
Sasi~Kiran Yelamarthi, Shiva~Krishna Reddy, Ashish Mishra, and Anurag Mittal.
\newblock A zero-shot framework for sketch based image retrieval.
\newblock In {\em Proceedings of the European Conference on Computer Vision
  (ECCV)}, pages 300--317, 2018.

\bibitem{yu2016SketchMeShoe}
Qian Yu, Feng Liu, Yi-Zhe Song, Tao Xiang, Timothy~M Hospedales, and
  Chen-Change Loy.
\newblock Sketch me that shoe.
\newblock In {\em Proc. CVPR}, pages 799--807, 2016.

\bibitem{Zhang2020SketchGCNZSSBIR}
Zhaolong Zhang, Yuejie Zhang, Rui Feng, Tao Zhang, and Weiguo Fan.
\newblock Zero-shot sketch-based image retrieval via graph convolution network.
\newblock In {\em Proceedings of the AAAI Conference on Artificial
  Intelligence}, volume~34, pages 12943--12950, 2020.

\end{thebibliography}


\begin{thebibliography}{10}\itemsep=-1pt

\bibitem{Caron2020SwavSelfS}
Mathilde Caron, Ishan Misra, Julien Mairal, Priya Goyal, Piotr Bojanowski, and
  Armand Joulin.
\newblock Unsupervised learning of visual features by contrasting cluster
  assignments.
\newblock {\em Advances in Neural Information Processing Systems},
  33:9912--9924, 2020.

\bibitem{Caron2021DINOSelfS}
Mathilde Caron, Hugo Touvron, Ishan Misra, Herv{\'e} J{\'e}gou, Julien Mairal,
  Piotr Bojanowski, and Armand Joulin.
\newblock Emerging properties in self-supervised vision transformers.
\newblock In {\em Proceedings of the IEEE/CVF International Conference on
  Computer Vision}, pages 9650--9660, 2021.

\bibitem{Dosovitskiy2021ViT}
Alexey Dosovitskiy, Lucas Beyer, Alexander Kolesnikov, Dirk Weissenborn,
  Xiaohua Zhai, Thomas Unterthiner, Mostafa Dehghani, Matthias Minderer, Georg
  Heigold, Sylvain Gelly, Jakob Uszkoreit, and Neil Houlsby.
\newblock An image is worth 16x16 words: Transformers for image recognition at
  scale.
\newblock In {\em International Conference on Learning Representations}, 2021.

\bibitem{resnet}
Kaiming He, Xiangyu Zhang, Shaoqing Ren, and Jian Sun.
\newblock Deep residual learning for image recognition.
\newblock In {\em Proc. CVPR}, pages 770--778, 2016.

\bibitem{Jegle2021PerceiverIO}
Andrew Jaegle, Sebastian Borgeaud, Jean-Baptiste Alayrac, Carl Doersch, Catalin
  Ionescu, David Ding, Skanda Koppula, Daniel Zoran, Andrew Brock, Evan
  Shelhamer, et~al.
\newblock Perceiver io: A general architecture for structured inputs \&
  outputs.
\newblock {\em arXiv preprint arXiv:2107.14795}, 2021.

\bibitem{Jaegle2021Perceiver}
Andrew Jaegle, Felix Gimeno, Andy Brock, Oriol Vinyals, Andrew Zisserman, and
  Joao Carreira.
\newblock Perceiver: General perception with iterative attention.
\newblock In {\em International conference on machine learning}, pages
  4651--4664. PMLR, 2021.

\bibitem{Shvetsova2022EverythingAtOnce}
Nina Shvetsova, Brian Chen, Andrew Rouditchenko, Samuel Thomas, Brian
  Kingsbury, Rogerio~S Feris, David Harwath, James Glass, and Hilde Kuehne.
\newblock Everything at once-multi-modal fusion transformer for video
  retrieval.
\newblock In {\em Proceedings of the IEEE/CVF Conference on Computer Vision and
  Pattern Recognition}, pages 20020--20029, 2022.

\bibitem{Tan2019efficientnet}
Mingxing Tan and Quoc Le.
\newblock Efficientnet: Rethinking model scaling for convolutional neural
  networks.
\newblock In {\em International conference on machine learning}, pages
  6105--6114. PMLR, 2019.

\bibitem{Tian2022TVTZSSBIR}
Jialin Tian, Xing Xu, Fumin Shen, Yang Yang, and Heng~Tao Shen.
\newblock Tvt: Three-way vision transformer through multi-modal hypersphere
  learning for zero-shot sketch-based image retrieval.
\newblock {\em Proceedings of the AAAI Conference on Artificial Intelligence},
  36(2):2370--2378, 2022.

\bibitem{Touvron2021DeiT}
Hugo Touvron, Matthieu Cord, Matthijs Douze, Francisco Massa, Alexandre
  Sablayrolles, and Herv{\'e} J{\'e}gou.
\newblock Training data-efficient image transformers \& distillation through
  attention.
\newblock In {\em International Conference on Machine Learning}, pages
  10347--10357. PMLR, 2021.

\bibitem{Yelamarthi2018FrameworkZSSBIR}
Sasi~Kiran Yelamarthi, Shiva~Krishna Reddy, Ashish Mishra, and Anurag Mittal.
\newblock A zero-shot framework for sketch based image retrieval.
\newblock In {\em Proceedings of the European Conference on Computer Vision
  (ECCV)}, pages 300--317, 2018.

\end{thebibliography}
}

\end{document}